# Web Document Categorization Using Naive Bayes Classifier and Latent Semantic Analysis

Alireza Saleh Sedghpour
School of Computer Engineering
Iran University of Science and Technology
Tehran, Iran
m_saleh@alumni.iust.ac.ir

Mohammad Reza Saleh Sedghpour
PhD Student at Department of Computing Science
Umeå University
Umeå, Sweden
msaleh@cs.umu.se

*Abstract—* **A rapid growth of web documents due to heavy use of World Wide Web necessitates efficient techniques to efficiently classify the document on the web. It is thus produced High volumes of data per second with high diversity. Automatically classification of these growing amounts of web document is One of the biggest challenges facing us today. Probabilistic classification algorithms such as Naive Bayes have become commonly used for web document classification. This problem is mainly because of the irrelatively high classification accuracy on plenty application areas as well as their lack of support to handle high dimensional and sparse data which is the exclusive characteristics of textual data representation. also it is common to Lack of attention and support the semantic relation between words using traditional feature selection method When dealing with the big data and large-scale web documents. In order to solve the problem, we proposed a method for web document classification that uses LSA to increase similarity of documents under the same class and improve the classification precision. Using this approach, we designed a faster and much accurate classifier for Web Documents. Experimental results have shown that using the mentioned preprocessing can improve accuracy and speed of Naive Bayes availably, the precision and recall metrics have indicated the improvement.**

## 1 Introduction

Today, with the advancement of many technologies including with the development of mobile World Wide Web, the usage of web content is rapidly increasing. The classification mapping algorithm between web page contents and predefined classes, been become ever more complicated in the volume and variety of web documents. Web documents classification plays a vital status on the Internet information management, convenient retrieval, web page crawling and user profile identification [1-2].

The classification techniques have been extensively used in different knowledge areas such as machine learning, pattern recognition, data mining, and information retrieval with practical applications in a number of various domains, such as e-marketing, image classification, system intrusion detection, healthcare monitoring, document categorization.

The central focus of this paper is on web document classification problem. Since the Most text data produced today, are in non-structural format
And Statistical learning techniques and machines cannot easily understand or analyze text documents. Also displays documents by terms (words and / or similar symbols) and duplicates usually .So, first, unstructured data should be converted into structure.Then, must be classified and structured using classification algorithms and special methods. Text classification is fraught with challenges, including high dimensionality of the feature space, where each unique word represents a feature [4]. Documents can be sparse with respect to the features when mapped into a structured format, to do this it is necessary to reduce the dimensions of the document.

These concepts have applications in a wide variety of areas in text mining. Some examples of them, consist of [5]: news filtering and organization, document organization and retrieval, opinion mining email classification and spam filtering, etc.

The definition of text classification is commonly stated as automatically organizing documents into predeter-mined categories (or classes). Several text classification algorithms depend on distance or similarity metrics which compare pairs of web documents. For this reason, similarity measures have central role in document categorization [6].

Text is usually associated with a wide range of unimportant or useless features. The first step is the feature extraction from entire corpus, further, preprocessing steps such as tokenization of text contents and removal of unnecessary tokens from the corpus is required. The processed tokens are used to build a vocabulary of the terms for the entire corpus. This is the main reason why feature extraction process is one of the difficult tasks in the problem of text classification. While a few research literatures have focused on this problem, the aim of this paper is to introduce a method for extracting the most relevant features and classification of text to improve the results.

The dimensionality reduction methods or feature selection algorithms are used to save both time and space for computation. Then, machine learning methods, like SVM [7], ANN [8], Rochio [9] and are applied for classifying web pages.

In this paper, firstly, the introduction and defining of the LSA model as a method for selecting the feature has been addressed. In this regard, we have segmented the relationships between traits (words) and categories And establish their production models according to different types of text data. In the next phase, The Bayes Naive classification is trained based on the finite-dimensional themed matrix, derived from the feature words derived from probabilistic distributions of LSA models. It combines the outstanding feature dimensionality reduction and text representation capabilities of LSA with the powerful classification ability of Naive Bayes to improve the text classification performance.

The rest of this paper is organized as follows. In Section 2, we introduce the Naive Bayes as text classification algorithm, and LSA as a feature selection method. In section 3, the proposed method is presented. The evaluation metric and experimental results have been discussed in Section 4. Finally in Section 5, the conclusions are discussed and future works are also explained.

## 2  Research topic literature

One of the most fundamental tasks that before any classification task, needs to be accomplished is that of document representation and feature selection. While feature selection is also desirable in other classification tasks, it is especially important in text classification due to the high dimensionality of text features and the existence of irrelevant (noisy) features. In this paper we use latent semantic analysis as s feature selection for Web document categorization (section 2.1). For classification task, we used Naïve Bayes (NB), which is explained in section 2.2.

### 2.1   Latent Semantic Analysis (LSA)

Latent semantic analysis (LSA) [10] is basically proposed as a feature selection method, which is widely used in text categorization. Once a term-by-document matrix is constructed, LSA requires the singular value decomposition of this matrix to construct a semantic vector space which can be used to represent conceptual term-document associations.

From the training documents, we can get the term by document matrix S(m × n), it means there are if we have m distinct terms in a n documents collection and m ≥ n. The singular value decomposition of S is defined as:

$$S = U \Sigma V^T \qquad (1)$$

Where U and V are the matrices of the term vectors and document vectors. $\Sigma = diag(\sigma_1, \ldots, \sigma_n)$ is the diagonal matrix of singular values.

In accordance with the formulas for reducing the dimensions, we can simply choose the $w$ largest singular values and the corresponding left and right singular vectors that shown in figure 1, the best approximation of S with rank-$f$ matrix calculated as follows:

$$S_f = U_f \Sigma_f V_f^T \qquad (2)$$

where $U_f$ is comprised of the first $f$ columns of the matrix U and $V_f^T$ is comprised the first $f$ rows of matrix $V^T$, $\Sigma_f = diag(\sigma_1, \ldots, \sigma_n)$ is the first $f$ factors, the ma-trix $S_f$ ignoring noise due to word choice and most of the important base structure that are in the association of terms and documents

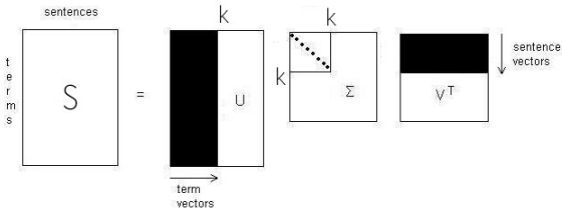

**Figure 1:** Singular value decomposition [10]

The explanation of applying the SVD to the terms by sentences matrix S can be made from two different opinion which are called Semantic Perspectives and transformation Perspectives. From semantic opinion, the SVD derives the latent semantic structure from the document represented by matrix S [11]. From transformation opinion, the SVD derives a mapping between the m-dimensional space spawned by the weighted term-frequency vectors and the r-dimensional singular vector space [11].

When LSA is used for feature selection, a query is represented as a vector in *f*-dimensional space. And then the query is compared to the documents. The query is represented by:

To select feature with LSA, First appears a query as a vector in the f-dimension space in the same way with the document collection, Then the query is compared to the records. The query is defined as follows:

$$\hat{q} = q^T U_f \Sigma_f^{-1} \qquad (3)$$

And each document is represented by:

$$\hat{d} = d^T U_f \Sigma_f^{-1} \qquad (4)$$

Once the query and the documents are represented, can be computed using cosine similarity coefficient, the relevance value between the query and documents those documents exceeding some cosine threshold are returned as relevance to the query [9].

## 2.2 What is the Naïve Bayes classifier?

Naïve Bayes classifier are a family of simple probabilistic classifiers based on applying Bayes' theorem. It has strong and easy independence assumptions between the features. J. Chen and al in their research, the description of the Naïve Bayes classifier [12] as follows:

There are two different models of Naïve Bayes classifiers in common that including The Multi-Variate Bernoulli Event Model and the Multinomial Event Model [13]. Both of these models use in the area of text classification. The first model are called the multi-variable Bernoulli event model and the second model is the polynomial event [13], both of them are based on Bayes' law to classify document categorization. Given a document $doc_i$, for each $doc_i$ document,

The probability of each class $cl_j$ is calculated as follows:

$$P(cl_j \vee doc_i) = \frac{P(doc_i | cl_j) . P(cl_j)}{P(doc_i)} \qquad (5)$$

As $P(doc_i)$ is the same for all class, then $label(doc_i)$, the class label of $doc_i$ can be determined by:

$$label(doc_i) = argMax_{coc_j}\{P(doc_i | cl_j) . P| \qquad (6)$$

The calculation of probability $P(doc_i | cl_j)$ in (6) is different in these two models. , a vocabulary V is given in the multi-variate Bernoulli event model. A document is represented with a vector of |V| dimensions. The *f*th dimension of the vector corresponds to word $w_f$ from V and is either 1 or 0, indicating whether word $w_f$ occurs in the document. To simplify the calculation of probability $P(doc_i | cl_j)$ in the Naïve Bayes assumption is made in this model: that in a document the probability of the occurrence of each word is independent of the occurrence of other words. Suppose document $d_i$ is represented with the vector ($t_1, t_2, …, t_{|V|}$), then $P(doc_i | cl_j)$ can be calculated under the Naïve Bayes assumption as:

$$P(doc_i | cl_j) = \prod_{f=1}^{V \vee} P(w_f | cl_j)^{t_f} (1 - P(w_f | cl_j))^{1-t_f} \square \qquad (7)$$

In the multinomial event model, a document is regarded as ''a bag of words''. No order of the words is considered, but the frequency of each word in the document is captured. In this model, a similar Naïve Bayes assumption is made: that the probability of the occurrence of each word in a document is independent of the word's position and the occurrence of other words in the document. Denote the number of times word $w_f$ occurs in document $d_i$ as $n_{if}$. Then the probability $P(doc_i|cl_j)$ from (6) can be computed by:

$$P(doc_i|cl_j) = P(|doc_i|)|doc_i|! \prod_{f=1}^{V} \frac{P(w_f|cl_j)^{n_{if}}}{n_{if}!} \quad (8)$$

Where $|d_i|$ is the number of words in document $|doc_i|$. Given a training set $Doc$, the probability $P(cl_j)$ from (6) is estimated as:

$$P(cl_j) = \frac{1+n_j}{1+n_{all}} \quad (9)$$

Where $n_j$ is the number of documents in class $c_j$, l is the number of classes, and $n_{all}$ is the number of all documents in the training set $D$. There are two ways to calculate probability $P(w_k|c_j)$ in (7) and (8). By the first means $P(w_k|c_j)$ is computed as:

$$P(w_k|c_j) = \frac{1+n_{c_j k}}{n_{all}+n_j} \quad (10)$$

where $n_j$ and $n_{all}$ is the same as that in (9), and $n_{c_j k}$ is the number of documents in class $c_j$ that contain word $w_k$. In this research, we will compute $P(w_k|c_j)$ in this way.

By the second means $P(w_k|c_j)$ is estimated as:

$$P(w_k|c_j) = \frac{1+N_{c_j k}}{N_{all}+N_j} \quad (11)$$

where $N_j$ is the number of words in class $c_j$, $N_{c_j k}$ is the number of word $w_k$ in class $c_j$, and $N_{all}$ is the number of all words in the training set $D$. As McCallum and Nigam [13] pointed out, the multinomial model performs usually better than the multi-variate Bernoulli model. So we use the former in this research.

## 3 Proposed Model for Web Document Categorization

This section proposes a method for categorization of Web Documents. Fig. 1 shows the proposed process.

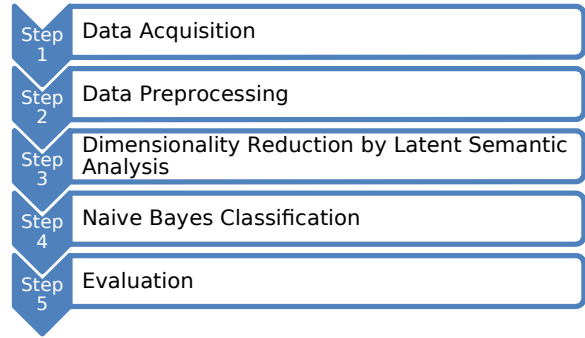

**Figure 2:** Proposed text classification process

In step1 the documents are collected using web searching by google search engine for six subjects that will be explained thoroughly in section 4.2.

In step2, preprocess of training and test data is done that includes removing stop words from text, outlier data detection and elimination and converting the texts to matrix using term frequency–inverse document frequency (TF-IDF) which is the most widely used term weight algorithm. Now there are two matrices, one for training and the other for test data. Both of these matrices are exposed to latent semantic analysis in step 3 to detect and eliminate unimportant features that leads to performance enhancement.

LSA is a two-stage process and includes learning and analysis of the indexed data [14]. During the learning phase LSA performs an automatic document indexing. The process starts with construction of a matrix A whose columns are associated with documents, and whose rows with terms (words or key-phrases). The matrix cell A($i$, $j$) contain the frequency (transformed using TF-IDF) of term $i$ in document $j$.

The second phase is the analysis. Most often this includes a study of the proximity between a couple of documents, a couple of words or between a word and a document. A simple mathematical transformation using the singular values and vectors from the training phase permits to obtain the vector for a non-indexed text. This allows the LSA for classification of text using Naïve Bayes model in step 4. The both of training and test matrices are used respectively for learning by Naïve Bayes algorithm and running on the test matrix to compute the performance measures of proposed method.

Finally in step 5, evaluation of proposed web document classification is performed in terms of precision and recall metrics.

# 4 Evaluation Results

This section introduces evaluation metrics to evaluate web document classifications and provides empirical evidence about performance of the proposed method for web document categorization.

## 4.1 Evaluation Metrics

Before proposing evaluation metrics, we need to introduce some notations. These notations are shown in table 1. We use the expert judgment for determining if a document belongs to class, or not. Table 1 shows relationship between the expert judgment and result of text classification method.

**TABLE 1:** Notations used in evaluation metrics

| Notation | Description |
|---|---|
| $FP_i$ | number of texts that are detected as positive for class $i$ incorrectly |
| $TP_i$ | number of texts that are detected as positive in class $i$ correctly |
| $FN_i$ | number of texts that are detected as negative in class $i$ incorrectly |
| $TN_i$ | number of texts that are detected as negative in class $i$ incorrectly |

Now we are ready to define our evaluation metrics which is defined in (12) and (13).

$$Pr = \frac{\sum_{i=1}^{n} Pr_i}{n} = \frac{\sum_{i=1}^{n}\left(\frac{|TP_i|}{(|TP_i|+|FP_i|)}\right)}{n} \quad (12)$$

$$\Re = \frac{\sum_{i=1}^{n} \Re_i}{n} = \frac{\sum_{i=1}^{n}\left(\frac{|TP_i|}{(|TP_i|+|FN_i|)}\right)}{n} \quad (13)$$

Where *Pr* and *Re* represent the precision and recall of web document classification. Both of these values are between 0 and 1 and values closer to 1 are better.

## 4.2 Experimental Results

The training data set we used for our experiment contained 1550 text documents. After preprocessing the data set, 2860 attributes (or terms) remained. This data set has collected around six subjects which are enumerated in table 2. Table 2 shows distribution of training documents in each of domains.

**TABLE 2:** Number of training documents in each of classes

| Number of docs | Class subject |
|---|---|
| 400 | Computer |
| 350 | Social |
| 150 | War |
| 300 | Political |
| 100 | Human Rights |
| 250 | Stock |

Documents obtained from step 1 of the process were preprocessed, outlier data were removed and TF-IDF matrix of data is prepared.

The test data contained 270 document and is collected and prepared in the same manner as train data. Table 3 shows distribution the training documents in each domains.

**TABLE 3:** Number of test documents in each of classes

| Number of docs | Class subject |
|---|---|
| 90 | Computer |
| 60 | Social |
| 20 | War |
| 50 | Political |
| 20 | Human Rights |
| 30 | Stock |

Fig. 2 shows the distributions of training and test data in each domain.

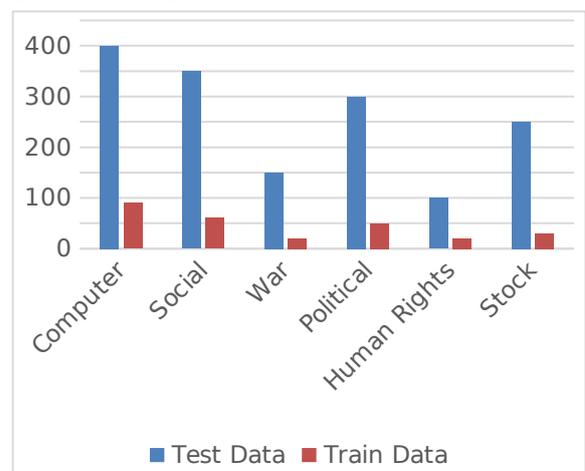

**Figure 2: Distributions of training and test data in each domain**

Table 4 shows assignment of documents to classes after classifying.

**TABLE 4:** Number of test documents in each of classes

| Number of true assignments | Number of assigned docs | Class subject |
|---|---|---|
| 90 | 91 | Computer |
| 50 | 58 | Social |
| 21 | 25 | War |
| 47 | 48 | Political |
| 12 | 16 | Human Rights |
| 32 | 32 | Stock |

Fig. 3 shows correct assignments of test documents to each class.

According to (12) and (13) we calculated the precision and recall of our document classification method and we yielded 91.5% for precision and 86.2% for recall. We have done separate experiment for classification of documents using only Naïve Bayes (without dimensionality reduction) to compare its result with our proposed method. The comparison of these two method are shown in table 5.

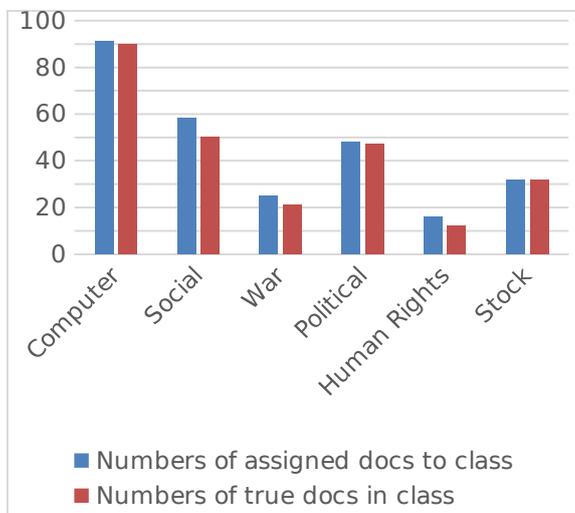

**Figure 3:** correct assignments of test documents to classes

**TABLE 5:** Comparison of precision and recall between Naïve Bayes and Latent Semantic Analysis- Naïve Bayes

| Precision | Recall | Experiment |
|---|---|---|
| 91.5% | 86.2% | Naïve Bayes-LSA |
| 84.1% | 74.1% | Naïve Bayes |
| **7.4%** | **12.1%** | **Improvement** |

As table 5 shows, we have improved precision and recall for 7.4 and 12.1 percent respectively.

## 5 Conclusion and Future Work

This paper presented a method for texts classification using latent semantic analysis and Naïve Bayes. Using Naïve Bayes for texts classification has reasonable performance but it can be improved. We applied latent semantic analysis for dimensionality reduction which is considered as feature selection and a part of overall preprocess of web documents to be classified. The experimental results showed that combing LSA and Naïve Bayes models improves precision and recall of web document classification compared to the only using Naïve Bayes.

Future work could include extending the proposed method for hierarchical web document classification. Another area of research related to this research is the adaptation of proposed method for distributed environments and Map Reduce programming paradigm.